\documentclass[11pt]{article}

\usepackage[preprint]{acl}

\usepackage{times}
\usepackage{latexsym}

\usepackage[T1]{fontenc}

\usepackage[utf8]{inputenc}

\usepackage{microtype}

\usepackage{inconsolata}

\usepackage{graphicx}

\usepackage[symbol]{footmisc}

\title{Breaking the Static Graph: Context-Aware Traversal for Robust Retrieval-Augmented Generation}

\author{
 \textbf{Kwun Hang Lau} \textsuperscript{1,2}\footnotemark[2],
 \textbf{Fangyuan Zhang} \textsuperscript{1},
 \textbf{Boyu Ruan} \textsuperscript{1},
 \\
 \textbf{Yingli Zhou}\textsuperscript{3},
 \textbf{Qintian Guo}\textsuperscript{2},
 \textbf{Ruiyuan Zhang}\textsuperscript{2},
 \textbf{Xiaofang Zhou}\textsuperscript{2}
\\
 \textsuperscript{1}Huawei Hong Kong Research Center, Hong Kong;\\
 \textsuperscript{2}The Hong Kong University of Science and Technology, Hong Kong;\\
 \textsuperscript{3}The Chinese University of Hong Kong, Shenzhen
}

\usepackage{amsmath}
\usepackage{booktabs}
\usepackage{multirow}
\usepackage{multicol}
\usepackage{amssymb}
\usepackage{adjustbox}
\usepackage{hyperref}

\begin{document}
\maketitle

\footnotetext[2]{Internship with Huawei Hong Kong Research Center.}
\renewcommand*{\thefootnote}{\arabic{footnote}}

\begin{abstract}
Recent advances in Retrieval-Augmented Generation (RAG) have shifted from simple vector similarity to structure-aware approaches like HippoRAG, which leverage Knowledge Graphs (KGs) and Personalized PageRank (PPR) to capture multi-hop dependencies. However, these methods suffer from a "Static Graph Fallacy": they rely on fixed transition probabilities determined during indexing. This rigidity ignores the query-dependent nature of edge relevance, causing semantic drift where random walks are diverted into high-degree "hub" nodes before reaching critical downstream evidence. Consequently, models often achieve high partial recall but fail to retrieve the complete evidence chain required for multi-hop queries. To address this, we propose CatRAG, Context-Aware Traversal for robust RAG, a framework that builds on the HippoRAG 2 architecture and transforms the static KG into a query-adaptive navigation structure. We introduce a multi-faceted framework to steer the random walk: \textbf{(1) Symbolic Anchoring}, which injects weak entity constraints to regularize the random walk; \textbf{(2) Query-Aware Dynamic Edge Weighting}, which dynamically modulates graph structure, to prune irrelevant paths while amplifying those aligned with the query's intent; and \textbf{(3) Key-Fact Passage Weight Enhancement}, cost-efficient bias that structurally anchors the random walk to likely evidence. Experiments across four multi-hop benchmarks demonstrate that CatRAG consistently outperforms state-of-the-art baselines. Our analysis reveals that while standard Recall metrics show modest gains, CatRAG achieves substantial improvements in reasoning completeness—the capacity to recover entire evidence path without gaps. These results reveal that our approach effectively bridges the gap between retrieving partial context and enabling fully grounded reasoning. Resources are available at  \href{https://github.com/kwunhang/CatRAG}{https://github.com/kwunhang/CatRAG}.
\end{abstract}

\section{Introduction}
Large Language Models (LLMs) have demonstrated transformative capabilities across a spectrum of natural language tasks, ranging from creative composition to complex code generation \cite{10.1145/3770084, li-etal-2024-fundamental, Ren_2024, touvron2023llama, brown2020languagemodelsfewshotlearners}. Despite these advances, the widespread deployment of LLMs still remains restricted by hallucinations\cite{xu2025hallucinationinevitable, liu2024surveyhallucination} in response generation, often caused by outdated training data or lack of domain-specific knowledge, resulting in seemingly plausible but actually incorrect content. Retrieval-Augmented Generation (RAG) \cite{gao2024retrievalaugmentedgenerationlargelanguage, 10.1145/3637528.3671470}  has emerged as a feasible solution to mitigate the issue, which incorporates external, reliable documents within LLM prompts for response generation.

Standard dense retrieval methods, which select document chunks based on semantic similarity \cite{izacard2022unsuperviseddenseinformationretrieval}, frequently fail in multi-hop reasoning scenarios when the answer relies on connecting disjoint facts. To overcome this limitation, recent research has shifted towards Structure-Aware RAG, which organizes information into hierarchical trees \cite{sarthi2024raptor} or global knowledge graphs \cite{guo2025lightrag} to capture long-range dependencies. Among these, the HippoRAG framework \cite{gutiérrez2024hipporag, gutiérrez2025hipporag2} distinguishes itself by leveraging Personalized PageRank (PPR) over Knowledge Graphs. HippoRAG simulates neurobiological memory mechanism, enabling deeper and more efficient knowledge integration that vector similarity alone cannot resolve.

However, a critical bottleneck remains in these graph-based paradigms: reliance on a static graph structure. In standard HippoRAG, the transition matrix guiding the Random Walk is fixed during indexing, determined solely by structural properties or a priori semantic similarity. This rigidity imposes two limitations. First, edge relevance is treated as context-independent. Consider the query: ``\textit{Which university did Marie Curie's doctoral advisor attend?}'' This requires a precise two-step traversal: \textit{Marie Curie} $\rightarrow$ \textit{Gabriel Lippmann} (Advisor) $\rightarrow$ \textit{École Normale Supérieure} (University). Yet, in a static graph, generic edges like \textit{Marie Curie} $\rightarrow$ \textit{Radioactivity} often possess dominant weights. Consequently, the random walk often suffers from semantic drift: it effectively retrieves the initial entity but is statistically diverted into irrelevant clusters before reaching the second-hop evidence. This results in a common failure mode where retrieval metrics (like Recall) appear high due to partial matches, yet the reasoning chain is broken.
Second, traversal is susceptible to the "hub node" problem, where high-degree entities (e.g., \textit{Nobel Prize}, \textit{French}) act as semantic sinks, disproportionately diluting the probability mass and causing the retrieval to drift into irrelevant documents.

To mitigate the constraints imposed by static graph topologies, we develop CatRAG (\textbf{C}ontext-\textbf{A}ware \textbf{T}raversal for robust \textbf{RAG}). This framework extends the HippoRAG 2 \cite{gutiérrez2025hipporag2} paradigm by integrating a novel optimization layer tailored for context-driven navigation. First, we introduce \textbf{Symbolic Anchoring}. By injecting explicitly recognized entities as weak topological anchors, we constrain the starting distribution to prevent immediate drift into generic hubs. Second, we introduce \textbf{Query-Aware Dynamic Edge Weighting}. By employing an LLM to assess the relevance of outgoing edges from seed entities, we dynamically modulate the graph edge weight, effectively pruning irrelevant paths while amplifying those aligned with the query's intent. Third, we propose \textbf{Key-Fact Passage Weight Enhancement}, a cost-efficient method to structurally anchor the random walk to documents containing verified evidentiary triples. It guides the random walk to documents that provide distinct evidence, rather than those containing only superficial mentions of seed entities.

We evaluate CatRAG across multiple multi-hop benchmarks. Results demonstrate that while our approach yields consistent gains in standard retrieval metrics, it achieves a significant breakthrough in reasoning sufficiency. CatRAG substantially improves Full Chain Retrieval—the ability to retrieve complete evidence chains—confirming that dynamic graph steering effectively mitigates semantic drift where static baselines fail.
\section{Related Work}
\subsection{Dense Retriever}
The foundational paradigm for RAG matches queries and documents in a shared vector space, evolving from probabilistic term-matching \cite{bm25} and dense bi-encoders \cite{contriever-2022} to granular late-interaction mechanisms \cite{santhanam2022colbertv2effectiveefficientretrieval}. Recently, the field has shifted toward Large Embedding Models like E5-Mistral \cite{wang2024improvingtextembeddingslarge}, NV-Embed \cite{lee2025nvembedimprovedtechniquestraining} and GritLM \cite{muennighoff2025GritLMgenerativerepresentationalinstructiontuning}, which repurpose LLMs to achieve superior benchmark performance \cite{muennighoff2022mteb}. However, these models remain constrained by the static nature of vector similarity. By compressing complex reasoning paths into a single geometrical proximity, they lack explicit multi-hop traversal mechanisms and frequently fail when queries and evidence are connected solely through intermediate bridge entities \cite{gutiérrez2024hipporag}.

\subsection{Structure-Aware RAG}
To transcend the limitations of flat vector spaces, recent works integrate explicit structural priors. Hierarchical approaches like RAPTOR \cite{sarthi2024raptor} organize text into recursive trees, while graph-based frameworks such as GraphRAG \cite{edge2025GraphRAG} and LightRAG \cite{guo2025lightrag} leverage Knowledge Graphs to traverse entity relationships. The state-of-the-art neuro-symbolic approach, HippoRAG \cite{gutiérrez2024hipporag} and its successor, HippoRAG 2 \cite{gutiérrez2025hipporag2}, simulates associative memory via PPR to link disparate facts. However, these methods suffer from the "Static Graph Fallacy": edge weights are fixed during indexing and cannot adapt to query-specific intent. This rigidity causes semantic drift, where high-degree "hub" nodes disproportionately dominate traversal probabilities, leading to the retrieval of structurally connected but contextually irrelevant paths.

\begin{figure*}[t]
  \includegraphics[width=\linewidth]{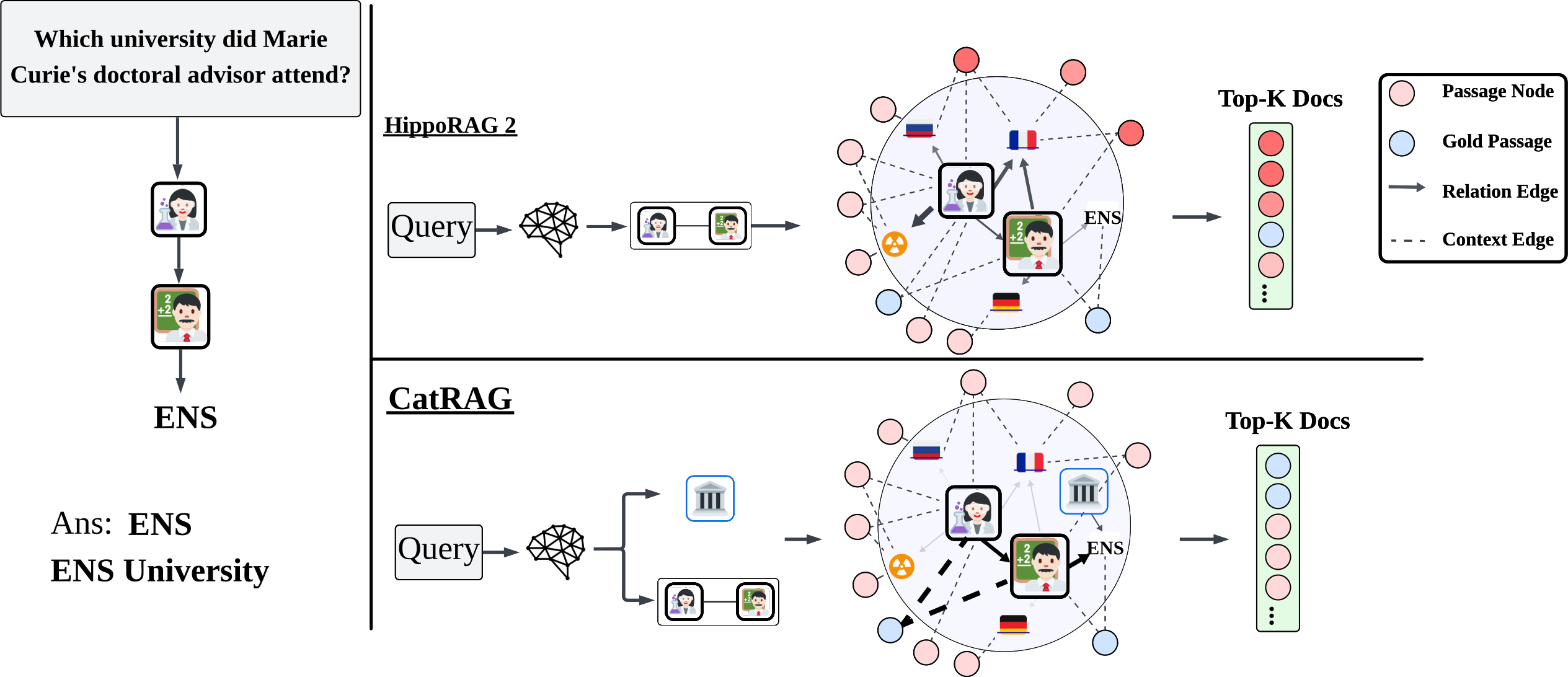}
  \caption{\textbf{
Comparison of graph traversal between HippoRAG 2 and CatRAG.} We illustrate the retrieval process for the multi-hop query ``Which university did Marie Curie's doctoral advisor attend?''. In HippoRAG 2 (top), the static graph structure causes semantic drift; probability mass is diverted to high-weight generic edges (e.g., \textit{Marie Curie}  $\rightarrow$ \textit{Radioactivity}), missing the downstream evidence \textit{ENS}. CatRAG (bottom) prevents this by applying (1) Symbolic Anchoring, injecting "University" as a weak seed, (2) Query-Aware Dynamic Edge Weighting amplifying relevant paths (e.g., \textit{Attend in ENS}) while pruning irrelevant ones, and (3) Key-Fact Passage Weight Enhancement to strength, boosting relevant context edge. This steers the random walk to successfully retrieve the complete evidence chain for \textit{ENS}.}
  \label{fig:Method show}
  \vspace{-10pt}
\end{figure*}

\subsection{Dynamic \& Adaptive Retrieval}
To address static retrieval limitations, iterative frameworks like IRCoT \cite{trivedi2023ircot} and Self-RAG \cite{asai2023SelfRaglearningretrievegenerate}, or agentic systems such as PRISM \cite{nahid2025PRISMagenticretrievalllms} and FAIR-RAG \cite{asl2025fairragfaithfuladaptiveiterative}, employ multi-step loops to refine search queries. While effective, these methods incur high latency and computational costs by requiring repeated LLM calls for multiple searches. CatRAG instead introduces a "one-shot" context-aware graph modification that dynamically re-weights edges before traversal. Unlike iterative cycles, our approach maintains the efficiency of a single retrieval pass, effectively combining the reasoning precision of adaptive methods with the speed and structural integrity of graph-based retrieval.




\section{Methodology}
\label{sec:methodology}
In this section, we propose three mechanisms to optimize HippoRAG 2's retrieval on a knowledge graph: \textit{Symbolic Anchoring}, \textit{Query-Aware Dynamic Edge Weighting} and \textit{Key-Fact Passage Weight Enhancement}, also present in Figure\ref{fig:Method show}. 

\subsection{Preliminaries}
We build our approach upon the graph structure defined in HippoRAG 2. The knowledge base is modeled as a directed graph $G = (V, E)$. The node set $V = V_{E} \cup V_{P}$ consists of entity phrases $V_{E}$ and passage nodes $V_{P}$.

The edge set $E$ is composed of three distinct types of semantic connections:
\begin{itemize}
    \item \textbf{Relation Edges ($E_{rel}$):} Edges between entity nodes ($u, v \in V_E$) derived from OpenIE triples. 
    \item \textbf{Synonym Edges ($E_{syn}$):} Edges connecting entity nodes with high vector similarity, capturing linguistic variations of the same concept.
    \item \textbf{Context Edges ($E_{ctx}$):} Edges linking a passage node $p \in V_P$ to the entity nodes $e \in V_E$ contained within it.
\end{itemize}

We adopt the Personalized PageRank (PPR) algorithm to model the retrieval process. The probability distribution over nodes at step $k$ is updated as:
\begin{equation}
\label{eq:ppr}
    \mathbf{v}^{(k+1)} = (1-d) \cdot \mathbf{e}_s + d \cdot \mathbf{v}^{(k)} \mathbf{T}
\end{equation}
where $\mathbf{e}_s$ is the personalized probability distribution over seed nodes, and $\mathbf{T}$ is the row-normalized transition matrix. In the standard framework, $\mathbf{T}$ is static. Our work focuses on dynamically refining $\mathbf{T}$ into a query-specific transition matrix $\hat{\mathbf{T}}_q$ to better capture the reasoning requirements of the user query.

\subsection{Symbolic Anchoring}
\label{subsec:symbolic_anchoring}
While the "Query to Triple" retrieval in HippoRAG 2 effectively captures implicit semantic cues, we argue that relying solely on dense vector alignment leaves the graph traversal susceptible to semantic drift. Without explicit constraints, the PPR propagation can easily be siphoned into high-degree "hub" nodes that have high similarity but lack precise relevance to the query. To mitigate this, we introduce \textbf{Symbolic Anchoring}, a regularization strategy that grounds the stochastic walk using explicit query constraints.

Rather than treating NER as an alternative retrieval path, we utilize extracted entities as strictly auxiliary \textit{topological anchors}. We extract a set of entities and inject them as weak seed, assigning reset probabilities for retrieval. We assign these symbolic anchors with small reset probabilities $\epsilon$, to ensure that their influence is subordinate to the initial entity from contextual triples.

This weak seeding serves a specific regulatory function: it aligns the PPR propagation with the query's intent. By placing a non-zero probability on the exact named entities mentioned in the query, we create a gravitational pull that resists the diffusion of probability mass into generic graph hubs. Even as the random walk explores the neighborhood defined by the static graph, these weak anchors ensure the traversal recurrently grounded to the specific entities in the query, effectively suppressing semantic drift. As a secondary benefit, this mechanism naturally balance the system's capability: it retains the triplet-based strength in interpreting implicit clues while ensuring robust coverage for containing explicit entity mentions.

\subsection{Query-Aware Dynamic Edge Weighting}
\label{subsec:dynamicWeight}
Current graph-based RAG models rely on a static transition matrix $T$, where transition probabilities are fixed during indexing. We argue that this rigidity induces stochastic drift: without query-specific guidance, the random walk indiscriminately diffuses probability mass into high-degree "hub" nodes that are structurally prominent but semantically irrelevant. To mitigate this, we approximate a query-conditional transition matrix $\hat{T}_q$, concentrating the random walk on edges that maximize information gain. We implement a \textbf{two-stage coarse-to-fine strategy} to dynamically modulate the weights of \textbf{relation edges} ($E_{rel}$).
\subsubsection{Adaptive Entity Contextualization}
To assist the LLM in evaluating the relevance of a transition from seed $u$ to neighbor $v$, we augment the prompt with a semantic summary of $v$. Since providing all connected facts for dense nodes is computationally intractable, we employ a conditional summarization strategy.
Let $\mathcal{F}(v)$ be the set of fact triples connected to entity node $v$. We define the context content $C(v)$ as:
\begin{equation}
    C(v) = 
    \begin{cases} 
    \text{Summary}(\mathcal{F}(v)) & \text{if } |\mathcal{F}(v)| > \tau \\
    \text{Concat}(\mathcal{F}(v)) & \text{otherwise}
    \end{cases}
\end{equation}
where $\tau$ is a density threshold. For information-dense nodes ($|\mathcal{F}(v)| > \tau$), we generate a concise summary; for sparse nodes, we use raw triples. This hybrid approach balances context completeness with token efficiency.

\subsubsection{Stage I: Coarse-Grained Candidate Pruning}
Evaluating the semantic relevance of every edge using an LLM is computationally prohibitive. Therefore, we first apply a topological filter to constrain the search space to the most plausible local neighborhoods.
We define two hyperparameters: the maximum number of seed entities $N_{seed}$ and the maximum number of edges per seed $K_{edge}$ for fine-grained alignment.  First, we select the top-$N_{seed}$ entity nodes based on their initial reset probabilities (derived from the dense retrieval alignment). Let $u$ be such a selected seed. 
For the seed phrase $u$ within top-$N_{seed}$, if the number of outgoing relation edges exceeds a threshold $K_{edge}$, we prune its outgoing edges by prioritizing the top-$K_{edge}$ neighbors based on the vector similarity between the query embedding and fact embeddings of relation edges.
Neighbors $v \notin \mathcal{N}_{top}(u)$ are bypassed by the scoring module and assigned a minimal \textit{Weak} weight. This step acts as a low-pass structural filter, discarding statistically improbable paths before the intensive semantic scoring.

\subsubsection{Stage II: Fine-Grained Semantic Probability Alignment}
In the second stage, we refine the weights of the surviving edges in $\mathcal{N}_{top}(u)$ to minimize semantic drift. While vector similarity (Stage I) captures general relatedness, it often fails to distinguish between generic associations and precise evidentiary links.
We employ a Large Language Model (LLM) as a discrete approximation of the conditional transition probability $P(v | u, q)$. The LLM evaluates the necessity of the transition $u \rightarrow v$ given the query $q$ and the neighbor's summary $C(v)$. We prompt the model to classify the relationship into discrete tiers $\mathcal{L} \in \{\text{Irrelevant, Weak, High, Direct}\}$. We define a mapping function $\phi: \mathcal{L} \rightarrow \mathbb{R}^+$ to project these judgments into scalar weights. The updated dynamic weight $\hat{w}_{uv}$ is computed as:
$$\hat{w}_{uv} = \phi\big(\text{LLM}(q, u, v, C(v))\big) \cdot w_{uv}^{(static)}$$
This modulation is asymmetric, applied only to forward edges originating from the seed set. By suppressing irrelevant edges and amplifying critical ones, we actively steer the PPR propagation, ensuring the traversal tunnels through the graph along the query's intent rather than diffusing into topological sinks.

\subsection{Key-Fact Passage Weight Enhancement}
\label{subsec:passageWeight}
In the directed graph setting, a seed entity node $u \in V_E$ may connect to multiple passage nodes $V_P$ via context edges. We aim to bias the walk towards passages containing "Key Facts"—fact triplets that were explicitly identified and filtered during the filtering Recognition memory filtering proposed in HippoRAG 2.

Let $\mathcal{T}_{seed}$ be the set of verified seed triples. We identify a "Key Fact" connection if the edge $E_{ctx}$ from seed entity $u$ to passage $p$ is supported by a triple in $\mathcal{T}_{seed}$. We enhance the weight of such edges:
\begin{equation}
    \hat{w}_{up} = w_{up} \cdot (1 + \beta \cdot \mathbb{I}(u, p \in \mathcal{T}_{seed}))
\end{equation}
where $\beta$ is a boost factor and $\mathbb{I}(\cdot)$ is an indicator function. 

This enhancement prioritizes passages providing evidentiary support. Unlike the previous module which requires LLM inference, the Key-Fact Enhancement is a purely algorithmic adjustment based on triple-matching. It incurs zero additional token cost and negligible latency, making it a highly efficient approach to guide the random walk.

\subsection{Unified Retrieval Process}
We integrate Symbolic Anchoring, Dynamic Edge Weighting, and Passage Enhancement to construct a query-adapted graph. Standard PPR (Eq. \ref{eq:ppr}) is executed on this refined structure. The resulting stationary distribution of PPR provides the final passage ranking, prioritizing nodes reachable via semantically relevant reasoning paths.

\section{Experimental Setup}
\subsection{Baselines}
We evaluate CatRAG against a comprehensive suite of baselines spanning two paradigms: standard RAG with retrieval methods, and structure-aware RAG. 

For standard retrieval comparisons, we employ several strong and widely used retrieval model,including \textbf{BM25} \cite{bm25}, \textbf{Contriever} \cite{contriever-2022}, \textbf{GTR} \cite{gtr-2022}, 
\textbf{text-embedding-3-small} \footnote{https://platform.openai.com/docs/models/text-embedding-3-small} model, to represent standard embedding-based approaches. 
Our primary comparison targets structure-aware RAG frameworks. We compare against \textbf{RAPTOR} \cite{sarthi2024raptor}, which constructs a recursive tree structure for hierarchical summarization, and \textbf{LightRAG} \cite{guo2025lightrag}, leverage a KG structure to generate corpus-level concept summaries. Crucially, our main baseline is \textbf{HippoRAG 2} \cite{gutiérrez2025hipporag2},the state-of-the-art in graph-based neuro-symbolic retrieval. We omit the original HippoRAG \cite{gutiérrez2024hipporag} from our evaluation, as HippoRAG 2 has demonstrated that it consistently outperforms its predecessor; thus, HippoRAG 2 serves as the most rigorous and relevant control. As CatRAG is built upon the HippoRAG 2 architecture, this comparison directly isolates the performance gains provided by our proposed methods.
\subsection{Datasets}
\begin{table}[t]
    \centering
    \caption{Dataset statistics.}
    \label{tab:dataset_statistics}
    \resizebox{\columnwidth}{!}{%
        \begin{tabular}{lrrrr}
            \toprule
            \textbf{Dataset} & \textbf{MuSiQue} & \textbf{2Wiki}  & \textbf{HotpotQA} & \textbf{HoVer} \\
            \midrule
            \textbf{\# of Queries} & 1,000 & 1,000 & 1,000 & 1,000 \\
            \textbf{\# of Passages} & 11,656  &  6,119 & 9,811 & 9,440\\
            \bottomrule
        \end{tabular}
    }
    \vspace{-5pt}
\end{table}

To evaluate the ability of \textit{CatRAG} to maintain precise retrieval in multi-hop scenarios, we conduct experiments on four benchmarks across two challenge types: \textbf{Multi-hop QA} and \textbf{Multi-hop Fact Verification}. We summarize the key statistics of these datasets in Table \ref{tab:dataset_statistics}.

\paragraph{\textbf{Multi-hop QA.}} We conduct experiments on \textbf{MuSiQue} \cite{trivedi2022musique}, \textbf{2WikiMultiHopQA} \cite{ho20202wikimultihop}, and \textbf{HotpotQA} \cite{yang2018hotpotqa}. These datasets require the system to reason over multiple passages to derive an answer. To ensure a fair comparison and reproducibility, we utilize the subsets defined in prior work \cite{gutiérrez2024hipporag}, which sampled 1,000 queries randomly and collected all candidate passages (including supporting and distractor passages) to form a corpus for each dataset. Crucially, \textbf{HotpotQA} and \textbf{2WikiMultiHopQA} are composed of 2-hop queries, while \textbf{MuSiQue} presents more challenging questions requiring 2 to 4 hops. 

\paragraph{\textbf{Multi-hop Fact Verification.}} We extend our evaluation to the \textbf{HoVer} dataset \cite{jiang2020hoverdataset} to test the robustness of our model in a claim verification setting. HoVer is adapted from HotpotQA but increases reasoning complexity by substituting named entities in the original claims with details from linked Wikipedia articles, thereby extending the reasoning chain to 3 and 4 hops. This substitution process creates deep, fragile reasoning chains where a single missed retrieval step results in failure. Following the protocol in HippoRAG, we randomly sample 1,000 claims from the dataset (specifically 3 and 4 hops) and form the retrieval corpus by collecting all candidate passages (supporting evidence and distractors) associated with the original lineage questions of selected claims.

\subsection{Metrics}
\label{metrics}
We report Recall@5 for standard retrieval evaluation and F1 for downstream QA. However, these aggregate metrics often mask incomplete reasoning, as models may retrieve partial evidence or guess correct answers without grounding. To rigorously assess reasoning integrity, we introduce \textbf{Full Chain Retrieval (FCR)}, defined as the percentage of queries where the retrieved context contains the \textit{entire} set of gold supporting documents. Furthermore, we report the \textbf{Joint Success Rate (JSR)}, which counts a query as successful only if the system achieves FCR \textit{and} the generated response contains the correct answer. This metric conceptually aligned with the strict evaluation established in the FEVER Shared Task \cite{thorne-etal-2018-FEVER} and HoVer \cite{jiang2020hoverdataset}, ensuring that accurate answer stem from complete evidentiary support rather than hallucinated or accidental correctness.

\begin{table*}[t!]
\centering
\small 
\setlength{\tabcolsep}{6pt} 
\caption{\textbf{Retrieval Performance (Recall@5).} 
Retrieval performance on multi-hop QA and fact verification datasets. LightRAG is not presented because it do not directly produce passage retrieval results.
}
\label{tab:retrieval_results}
\begin{tabular}{l|ccc|c}
\toprule
\textbf{Method} & \textbf{MuSiQue} & \textbf{2Wiki} & \textbf{HotpotQA} & \textbf{HoVer} \\
\midrule
\multicolumn{5}{l}{\textit{Standard Retrieval}} \\
BM25 & 31.6 & 52.1 & 64.8 & 50.8 \\
Contriever & 46.6 & 57.5 & 75.3 & 62.3 \\
GTR (T5-base) & 49.1 & 67.9 & 73.9 & 55.6 \\
text-embedding-3-small & 55.4 & 70.8 & 81.3 & 65.7 \\
\midrule
\multicolumn{5}{l}{\textit{Structure-Aware RAG}} \\
RAPTOR & 53.3 & 69.8 & 79.5 & 62.4  \\
HippoRAG 2 & \underline{61.4} & \underline{85.9} & \underline{87.1} & \underline{71.2} \\
\midrule
\textbf{CatRAG} & \textbf{64.9} & \textbf{87.0} & \textbf{89.5} & \textbf{76.8} \\ 
\bottomrule
\end{tabular}
\vspace{-3pt}
\end{table*}


\begin{table*}[t!]
\centering
\small
\setlength{\tabcolsep}{7pt}
\caption{\textbf{Downstream QA Performance.} QA performance on multi-hop QA and fact verification datasets using Llama-3.3-70B-Instruct as the QA reader. We report F1 for QA datasets, accuracy for the HoVer dataset. * denotes the results from \cite{gutiérrez2025hipporag2}.}
\label{tab:qa_results}
\begin{tabular}{l|ccc|c}
\toprule
\textbf{Method} & \textbf{MuSiQue} & \textbf{2Wiki} & \textbf{HotpotQA} & \textbf{HoVer} \\
\midrule
\multicolumn{5}{l}{\textit{Standard Retrieval}} \\
None & 26.1* & 42.8* & 47.3* & $-$ \\
BM25 & 22.9 & 39.9 & 54.1 & 61.4  \\
Contriever & 31.3 &  41.9 & 62.3 & 66.0 \\
GTR (T5-base) & 34.6 & 52.8 & 62.8 & 62.7 \\
text-embedding-3-small &  36.1 &  56.9 &  64.6 & 64.2 \\
\midrule
\multicolumn{5}{l}{\textit{Structure-Aware RAG}} \\
RAPTOR & 36.0 & 56.7 & 64.4 & 65.3 \\
LightRAG & 43.0 & 49.7 &  68.3 & 66.5 \\
HippoRAG 2 & \underline{43.2} & \underline{68.1} & \underline{69.4} & \underline{67.2} \\
\midrule
\textbf{CatRAG} & \textbf{45.0} & \textbf{69.7} & \textbf{71.4} & \textbf{69.0} \\
\bottomrule
\end{tabular}
\vspace{-5pt}
\end{table*}

\subsection{Implementation Details}
We implement CatRAG upon the HippoRAG 2 architecture, using GPT-4o-mini\footnote{https://platform.openai.com/docs/models/gpt-4o-mini} as the backbone for all LLM components and text-embedding-3-small as the retriever. While newer open-weight models like NV-Embed-v2 \cite{lee2025nvembedimprovedtechniquestraining} show strong performance, our primary objective is to isolate the topological gains provided by the CatRAG mechanism from the raw semantic capacity of the underlying encoder. For fair comparison, all structure-augmented baselines are reproduced using the same extractor and retriever. Downstream responses are generated by Llama-3.3-70B-Instruct using the top-5 retrieved passages. Key hyperparameters include: symbolic anchor reset probability $\epsilon = 0.2$ (weighted by inverse passage count $|P_i|^{-1}$), boost factor  $\beta=2.5$, dynamic weighting limits  $N_{seed}=5$ and $K_{edge}=15$. More implementation details and hyperparameters are provided in Appendix \ref{app:implementation}.

\begin{table*}[t!]
\centering
\small 
\setlength{\tabcolsep}{6pt} 
\caption{\textbf{Reasoning Completeness Evaluation.} We evaluation the \textbf{Full Chain Retrieval} (\textbf{FCR}) and \textbf{Joint Success Rate} (\textbf{JSR}) on multi-hop QA and fact verification datasets. LightRAG is not presented because it do not directly produce passage retrieval results.}
\label{tab:retrieval_completeness}
\begin{tabular}{l|ccc|c}
\toprule
\textbf{Method} & \textbf{MuSiQue} & \textbf{2Wiki} & \textbf{HotpotQA} & \textbf{HoVer} \\
\midrule
\multicolumn{5}{l}{\textit{Standard Retrieval}} \\
BM25 & 6.4/4.5 &  20.5/19.1 & 38.3/26.1 & 8.2/6.3 \\
Contriever & 11.3/8.3 & 27.2/23.9 & 54.1/37.6 & 18.1/13.8 \\
GTR (T5-base) & 15.0/11.5 & 35.8/31.3 & 53.0/37.9 & 10.3/6.8 \\
text-embedding-3-small & 21.1/13.8 & 41.6/34.9 & 64.9/46.3 & 22.1/16.0 \\
\midrule
\multicolumn{5}{l}{\textit{Structure-Aware RAG}} \\
RAPTOR & 19.6/13.2 & 40.1/34.2 & 61.4/44.2 & 18.6/13.7 \\
HippoRAG 2 & \underline{30.5/21.5} & \underline{66.1/53.0} & \underline{75.5/53.4} & \underline{34.8/26.2} \\
\midrule
\
\textbf{CatRAG} & \textbf{34.6/24.3} & \textbf{67.6/55.0} & \textbf{80.4/56.8} & \textbf{42.5/31.1} \\ 
\bottomrule
\end{tabular}
\vspace{-5pt}
\end{table*}
\section{Results}
\label{sec:results}
\paragraph{Standard Retrieval and QA.}
Table~\ref{tab:retrieval_results} and Table~\ref{tab:qa_results} demonstrate that CatRAG consistently outperforms all baselines across standard metrics. On the complex MuSiQue dataset (2--4 hops), CatRAG achieves a Recall@5 of 64.9\%, surpassing the dense retriever text-embedding-3-small by a substantial 8.1\% margin and confirming the necessity of structure-aware methods.\\
Compared to the state-of-the-art static baseline, HippoRAG 2 across all benchmarks, CatRAG raises Recall@5 to 89.5\% on HotpotQA and 76.8\% on HoVer.
This retrieval quality directly translates to downstream performance, where CatRAG yields the highest F1 scores across all datasets (e.g., 45.0\% on MuSiQue), validating that query-conditional edge weighting surfaces relevant evidence without disrupting structural integrity.

\paragraph{Strict Reasoning Completeness Evaluation.}
While standard metrics indicate general relevance, they often mask a critical failure mode in multi-hop retrieval: the loss of intermediate "bridge" documents that connect disjoint facts. To assess the recovery of the full evidence paths, we evaluate FCR and JSR in Table~\ref{tab:retrieval_completeness}.
CatRAG effectively mitigates probability dilution, achieving  an FCR of 34.6\% compared to 30.5\% for HippoRAG 2. The gain is most pronounced on HoVer, where precise 3--4 hop claim verification is required. CatRAG improves JSR to 31.1\%, a relative gain of 18.7\% over the HippoRAG2.
These results confirm that our dynamic steering successfully anchors the traversal to the specific bridge documents required for grounded reasoning.

\subsection{Ablation Study}
We conduct an ablation experiment, to isolate the contributions of Symbolic Anchoring, Query-Aware Dynamic Edge Weighting ($E_{rel}$), and Key-Fact Passage Weight Enhancement, with results summarized in Table~\ref{tab:ablation}. 
First, the removal of Symbolic Anchoring precipitates a consistent performance degradation, most notably a 3.2\% drop on HoVer. This confirms that injecting extracted entities as weak topological anchors is critical for mitigating semantic drift.
Second, excluding $E_{rel}$ weighting results in significant losses across all benchmarks, confirming that dynamically pruning irrelevant semantic branches is foundational to mitigating drift. 
Finally, we observe that Key-Fact Enhancement provides consistent gains across unstructured datasets (HotpotQA, MuSiQue, HoVer) where evidence is buried in dense text. On the highly structured dataset 2WikiMultiHopQA, this heuristic introduces slight noise, leading to a minor performance regression . However, given that real-world RAG scenarios involve messy, unstructured corpora, we prioritize the gains on the unstructured datasets. 
\begin{table}[t!]
\centering
\small
\caption{\textbf{Ablations.} We report passage recall@5 on multi-hop benchmarks using several alternatives to our final design in dynamic update.}
\resizebox{\columnwidth}{!}{%
    \begin{tabular}{lcccc}
    \toprule
     & MuSiQue & 2Wiki & HotpotQA & HoVer  \\
    \midrule
    CatRAG &  \textbf{64.9} & 87.0 & \textbf{89.5} & \textbf{76.8}  \\
    \midrule
    w/o Symbolic anchor& 63.0 & 86.1 & 88.6 & 73.6\\ 
     w/o$E_{rel}$ weighting& 63.2 & 85.6 &88.1 &75.0\\
     w/o Passage Enhance& 64.7 & \textbf{88.4} & 89.0 & 76.6 \\
    \bottomrule
    \end{tabular}
}
\label{tab:ablation}
\vspace{-5pt}
\end{table}
\section{Discussion}
\begin{figure}[t]
    \centering
    \includegraphics[width=\linewidth]{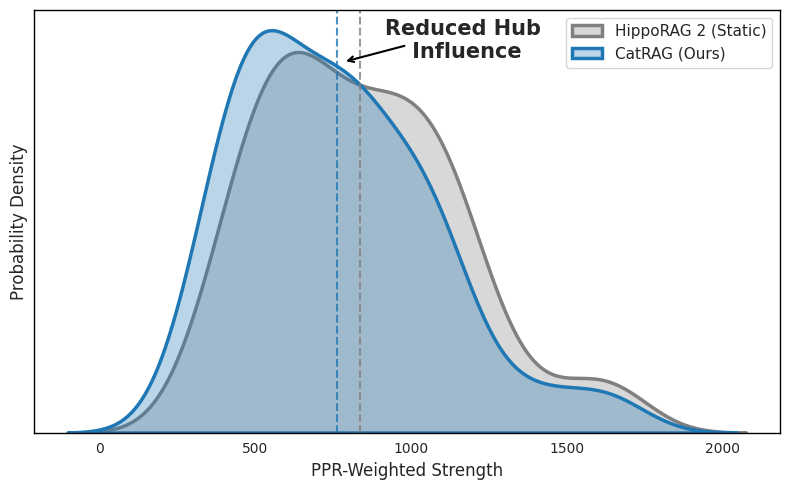}
    \caption{
        \textbf{Distribution of PPR-Weighted Node Strength ($\mathcal{S}_{ppr}$).} 
        Comparison of the HippoRAG 2 versus CatRAG. 
        The distribution for CatRAG is shifted to the left, indicating a reduction in the retrieval of high-degree "Hub" nodes. 
        The dashed lines represent the mean $\mathcal{S}_{ppr}$ for each method.
    }
    \label{fig:hub_bias}
    \vspace{-10pt}
\end{figure}
\subsection{Impact on Multi-Hop Dependency: Mitigating Hub Bias}
\label{sec:discussion_hub_bias}
A fundamental limitation of static graph retrieval is Hub Bias (or degree centrality bias). In standard formulations like HippoRAG 2, transition probabilities are determined by static structural properties. Consequently, random walks disproportionately converge on high-degree nodes (e.g., generic entities like "United States" or "Song"), which act as "topological sinks". We hypothesize that this structural noise disrupts multi-hop dependency by diverting the retrieval path away from the specific "bridge" entities required to connect disjoint facts.
\paragraph{Quantifying Semantic Drift.}
To assess whether our proposed framework mitigates this drift, we analyzed the topological properties of the top-10 retrieved entity nodes after PPR across 100 randomly sampled queries from MuSiQue. We introduce \textit{PPR-Weighted Strength} ($\mathcal{S}_{ppr}$) to measure the effective structural prominence of the retrieved context:
\begin{equation}
    \mathcal{S}_{ppr}(q) = \sum_{v \in \mathcal{V}_{top}} \hat{p}(v) \cdot \text{Strength}(v)
\end{equation}
where $\hat{p}(v)$ is the PPR probability mass of node $v$ re-normalized over the retrieved set $\mathcal{V}_{top}$ (i.e., $\sum \hat{p}(v) = 1$), and $\text{Strength}(v)$ is the weighted degree of the node. A higher $\mathcal{S}_{ppr}$ indicates that the PPR result is more reliant on generic, high-connectivity nodes.
\paragraph{Mitigation of Hub Bias.}
As illustrated in Figure~\ref{fig:hub_bias}, CatRAG exhibits a systematic structural shift toward specificity. The distribution of PPR-Weighted Strength for CatRAG is distinctively shifted to the left compared to the static baseline HippoRAG. CatRAG reduces the Mean PPR-Weighted Strength from 837.0 to 761.7. Furthermore, we quantified the probability mass allocated to "Super Hubs" (nodes in the top 1\% of weighted degree). While the baseline allocates 45.7\% of its probability mass to these generic hubs, our method significantly reduces this to 42.5\%.
\paragraph{Correlation with Reasoning Completeness.}
This structural correction directly explains the improvements in reasoning integrity observed in Table~\ref{tab:retrieval_completeness}. While the relative reduction in hub mass (7\%) may appear moderate, it represents a critical redistribution of probability mass away from topological distractors and toward specific bridge entities. This aligns with our results on the HoVer dataset, where avoiding generic associations is crucial for verification; specifically, this structural enhancement enables the 11\% relative improvement in JSR. By structurally decoupling prominence from relevance, CatRAG ensures that the retrieved context preserves the complete dependency chain, bridging the gap between partial recall and grounded reasoning.

\section{Conclusion}
\label{sec:conclusion}
We identify and address the "Static Graph Fallacy" inherent in current structure-aware RAG systems, where fixed transition probabilities predispose retrieval to semantic drift and prevent the recovery of complete evidence chains. We propose CatRAG, a framework that transforms the Knowledge Graph Traversal into a context-aware navigation structure. Experiment across multi-hop benchmarks demonstrate that CatRAG consistently outperforms baselines, including HippoRAG 2, while significantly reducing the bias of high-degree hub nodes. Our analysis reveals that these topological adjustments yield substantial improvements in reasoning completeness, effectively bridging the gap between retrieving partial context and enabling fully grounded, multi-hop reasoning.

\section*{Limitations}
While CatRAG significantly enhances reasoning completeness, it introduces certain trade-offs regarding efficiency. First, the mechanism for query-aware dynamic edge weighting requires run-time LLM inference to assess semantic relevance, which incurs additional computational overhead and latency compared to purely static graph traversals. Although we mitigate this via coarse-grained pruning, the approach remains more computationally intensive than standard dense retrieval. Furthermore, our experimental evaluation intentionally utilized standard embedding models (text-embedding-3-small) rather than larger, state-of-the-art embedding models to strictly isolate the topological gains provided by our framework from the raw semantic capacity of the encoder. Consequently, while our results demonstrate the superiority of dynamic traversal, the absolute performance ceiling of CatRAG could potentially be further elevated by integrating these larger foundational models in future work.
Due to proprietary data protection policies, the full source code cannot be publicly released. To mitigate this, we have provided full hyperparameter tables in to facilitate reimplementation.
\section{Ethical considerations}
This study utilizes four publicly available benchmark datasets, MuSiQue, 2WikiMultiHopQA, HotpotQA, and HoVer, which are standard in the field. These datasets are derived from Wikipedia/Wikidata sources and may therefore contain publicly available information about real people and may incidentally include sensitive topics; however, we did not collect new personal data or interact with human participants. Regarding computational resources and model access, we utilized GPT-4o mini and text-embedding-3-small via the Microsoft Azure API, and accessed Llama-3.3-70B-Instruct through the OpenRouter API. \\
According with AI Assistance policies, we acknowledge that we used generative AI tools to assist with code implementation and language polishing. All scientific content and results were verified by the authors.

\bibliography{custom}

@InProceedings{bm25,
author="Robertson, S. E.
and Walker, S.",
editor="Croft, Bruce W.
and van Rijsbergen, C. J.",
title="Some Simple Effective Approximations to the 2-Poisson Model for Probabilistic Weighted Retrieval",
booktitle="SIGIR '94",
year="1994",
publisher="Springer London",
address="London",
pages="232--241",
isbn="978-1-4471-2099-5"
}

@misc{contriever-2022,
      title={Unsupervised Dense Information Retrieval with Contrastive Learning}, 
      author={Gautier Izacard and Mathilde Caron and Lucas Hosseini and Sebastian Riedel and Piotr Bojanowski and Armand Joulin and Edouard Grave},
      year={2022},
      eprint={2112.09118},
      archivePrefix={arXiv},
      primaryClass={cs.IR},
      url={https://arxiv.org/abs/2112.09118}, 
}

@inproceedings{gtr-2022,
    title = "Large Dual Encoders Are Generalizable Retrievers",
    author = "Ni, Jianmo  and
      Qu, Chen  and
      Lu, Jing  and
      Dai, Zhuyun  and
      Hernandez Abrego, Gustavo  and
      Ma, Ji  and
      Zhao, Vincent  and
      Luan, Yi  and
      Hall, Keith  and
      Chang, Ming-Wei  and
      Yang, Yinfei",
    editor = "Goldberg, Yoav  and
      Kozareva, Zornitsa  and
      Zhang, Yue",
    booktitle = "Proceedings of the 2022 Conference on Empirical Methods in Natural Language Processing",
    month = dec,
    year = "2022",
    address = "Abu Dhabi, United Arab Emirates",
    publisher = "Association for Computational Linguistics",
    url = "https://aclanthology.org/2022.emnlp-main.669/",
    doi = "10.18653/v1/2022.emnlp-main.669",
    pages = "9844--9855"
    
}

@inproceedings{
sarthi2024raptor,
title={{RAPTOR}: Recursive Abstractive Processing for Tree-Organized Retrieval},
author={Parth Sarthi and Salman Abdullah and Aditi Tuli and Shubh Khanna and Anna Goldie and Christopher D Manning},
booktitle={The Twelfth International Conference on Learning Representations},
year={2024},
url={https://openreview.net/forum?id=GN921JHCRw}
}

@misc{guo2025lightrag,
      title={LightRAG: Simple and Fast Retrieval-Augmented Generation}, 
      author={Zirui Guo and Lianghao Xia and Yanhua Yu and Tu Ao and Chao Huang},
      year={2025},
      eprint={2410.05779},
      archivePrefix={arXiv},
      primaryClass={cs.IR},
      url={https://arxiv.org/abs/2410.05779}, 
}

@misc{gutiérrez2024hipporag,
      title={HippoRAG: Neurobiologically Inspired Long-Term Memory for Large Language Models}, 
      author={Bernal Jiménez Gutiérrez and Yiheng Shu and Yu Gu and Michihiro Yasunaga and Yu Su},
      year={2024},
      eprint={2405.14831},
      archivePrefix={arXiv},
      primaryClass={cs.CL},
      url={https://arxiv.org/abs/2405.14831}, 
}

@misc{gutiérrez2025hipporag2,
      title={From RAG to Memory: Non-Parametric Continual Learning for Large Language Models}, 
      author={Bernal Jiménez Gutiérrez and Yiheng Shu and Weijian Qi and Sizhe Zhou and Yu Su},
      year={2025},
      eprint={2502.14802},
      archivePrefix={arXiv},
      primaryClass={cs.CL},
      url={https://arxiv.org/abs/2502.14802}, 
}

@misc{trivedi2022musique,
      title={MuSiQue: Multihop Questions via Single-hop Question Composition}, 
      author={Harsh Trivedi and Niranjan Balasubramanian and Tushar Khot and Ashish Sabharwal},
      year={2022},
      eprint={2108.00573},
      archivePrefix={arXiv},
      primaryClass={cs.CL},
      url={https://arxiv.org/abs/2108.00573}, 
}

@misc{yang2018hotpotqa,
      title={HotpotQA: A Dataset for Diverse, Explainable Multi-hop Question Answering}, 
      author={Zhilin Yang and Peng Qi and Saizheng Zhang and Yoshua Bengio and William W. Cohen and Ruslan Salakhutdinov and Christopher D. Manning},
      year={2018},
      eprint={1809.09600},
      archivePrefix={arXiv},
      primaryClass={cs.CL},
      url={https://arxiv.org/abs/1809.09600}, 
}

@misc{ho20202wikimultihop,
      title={Constructing A Multi-hop QA Dataset for Comprehensive Evaluation of Reasoning Steps}, 
      author={Xanh Ho and Anh-Khoa Duong Nguyen and Saku Sugawara and Akiko Aizawa},
      year={2020},
      eprint={2011.01060},
      archivePrefix={arXiv},
      primaryClass={cs.CL},
      url={https://arxiv.org/abs/2011.01060}, 
}

@misc{jiang2020hoverdataset,
      title={HoVer: A Dataset for Many-Hop Fact Extraction And Claim Verification}, 
      author={Yichen Jiang and Shikha Bordia and Zheng Zhong and Charles Dognin and Maneesh Singh and Mohit Bansal},
      year={2020},
      eprint={2011.03088},
      archivePrefix={arXiv},
      primaryClass={cs.CL},
      url={https://arxiv.org/abs/2011.03088}, 
}

@inproceedings{li-etal-2024-fundamental,
    title = "Fundamental Capabilities of Large Language Models and their Applications in Domain Scenarios: A Survey",
    author = "Li, Jiawei  and
      Yang, Yizhe  and
      Bai, Yu  and
      Zhou, Xiaofeng  and
      Li, Yinghao  and
      Sun, Huashan  and
      Liu, Yuhang  and
      Si, Xingpeng  and
      Ye, Yuhao  and
      Wu, Yixiao  and
      Lin, Yiguan  and
      Xu, Bin  and
      Ren, Bowen  and
      Feng, Chong  and
      Gao, Yang  and
      Huang, Heyan",
    editor = "Ku, Lun-Wei  and
      Martins, Andre  and
      Srikumar, Vivek",
    booktitle = "Proceedings of the 62nd Annual Meeting of the Association for Computational Linguistics (Volume 1: Long Papers)",
    month = aug,
    year = "2024",
    address = "Bangkok, Thailand",
    publisher = "Association for Computational Linguistics",
    url = "https://aclanthology.org/2024.acl-long.599/",
    doi = "10.18653/v1/2024.acl-long.599",
    pages = "11116--11141"
}

@inproceedings{Ren_2024, series={KDD ’24},
   title={A Survey of Large Language Models for Graphs},
   url={http://dx.doi.org/10.1145/3637528.3671460},
   DOI={10.1145/3637528.3671460},
   booktitle={Proceedings of the 30th ACM SIGKDD Conference on Knowledge Discovery and Data Mining},
   publisher={ACM},
   author={Ren, Xubin and Tang, Jiabin and Yin, Dawei and Chawla, Nitesh and Huang, Chao},
   year={2024},
   month=aug, pages={6616–6626},
   collection={KDD ’24} }

@misc{brown2020languagemodelsfewshotlearners,
      title={Language Models are Few-Shot Learners}, 
      author={Tom B. Brown and Benjamin Mann and Nick Ryder and Melanie Subbiah and Jared Kaplan and Prafulla Dhariwal and Arvind Neelakantan and Pranav Shyam and Girish Sastry and Amanda Askell and Sandhini Agarwal and Ariel Herbert-Voss and Gretchen Krueger and Tom Henighan and Rewon Child and Aditya Ramesh and Daniel M. Ziegler and Jeffrey Wu and Clemens Winter and Christopher Hesse and Mark Chen and Eric Sigler and Mateusz Litwin and Scott Gray and Benjamin Chess and Jack Clark and Christopher Berner and Sam McCandlish and Alec Radford and Ilya Sutskever and Dario Amodei},
      year={2020},
      eprint={2005.14165},
      archivePrefix={arXiv},
      primaryClass={cs.CL},
      url={https://arxiv.org/abs/2005.14165}, 
}

@misc{touvron2023llama,
      title={LLaMA: Open and Efficient Foundation Language Models}, 
      author={Hugo Touvron and Thibaut Lavril and Gautier Izacard and Xavier Martinet and Marie-Anne Lachaux and Timothée Lacroix and Baptiste Rozière and Naman Goyal and Eric Hambro and Faisal Azhar and Aurelien Rodriguez and Armand Joulin and Edouard Grave and Guillaume Lample},
      year={2023},
      eprint={2302.13971},
      archivePrefix={arXiv},
      primaryClass={cs.CL},
      url={https://arxiv.org/abs/2302.13971}, 
}

@article{10.1145/3770084,
author = {Joel, Sathvik and Wu, Jie and Fard, Fatemeh},
title = {A Survey on LLM-based Code Generation for Low-Resource and Domain-Specific Programming Languages},
year = {2025},
publisher = {Association for Computing Machinery},
address = {New York, NY, USA},
issn = {1049-331X},
url = {https://doi.org/10.1145/3770084},
doi = {10.1145/3770084},
note = {Just Accepted},
journal = {ACM Trans. Softw. Eng. Methodol.},
month = oct
}

@misc{xu2025hallucinationinevitable,
      title={Hallucination is Inevitable: An Innate Limitation of Large Language Models}, 
      author={Ziwei Xu and Sanjay Jain and Mohan Kankanhalli},
      year={2025},
      eprint={2401.11817},
      archivePrefix={arXiv},
      primaryClass={cs.CL},
      url={https://arxiv.org/abs/2401.11817}, 
}

@misc{liu2024surveyhallucination,
      title={A Survey on Hallucination in Large Vision-Language Models}, 
      author={Hanchao Liu and Wenyuan Xue and Yifei Chen and Dapeng Chen and Xiutian Zhao and Ke Wang and Liping Hou and Rongjun Li and Wei Peng},
      year={2024},
      eprint={2402.00253},
      archivePrefix={arXiv},
      primaryClass={cs.CV},
      url={https://arxiv.org/abs/2402.00253}, 
}

@misc{gao2024retrievalaugmentedgenerationlargelanguage,
      title={Retrieval-Augmented Generation for Large Language Models: A Survey}, 
      author={Yunfan Gao and Yun Xiong and Xinyu Gao and Kangxiang Jia and Jinliu Pan and Yuxi Bi and Yi Dai and Jiawei Sun and Meng Wang and Haofen Wang},
      year={2024},
      eprint={2312.10997},
      archivePrefix={arXiv},
      primaryClass={cs.CL},
      url={https://arxiv.org/abs/2312.10997}, 
}

@inproceedings{10.1145/3637528.3671470,
author = {Fan, Wenqi and Ding, Yujuan and Ning, Liangbo and Wang, Shijie and Li, Hengyun and Yin, Dawei and Chua, Tat-Seng and Li, Qing},
title = {A Survey on RAG Meeting LLMs: Towards Retrieval-Augmented Large Language Models},
year = {2024},
isbn = {9798400704901},
publisher = {Association for Computing Machinery},
address = {New York, NY, USA},
url = {https://doi.org/10.1145/3637528.3671470},
doi = {10.1145/3637528.3671470},
booktitle = {Proceedings of the 30th ACM SIGKDD Conference on Knowledge Discovery and Data Mining},
pages = {6491–6501},
numpages = {11},
location = {Barcelona, Spain},
series = {KDD '24}
}

@misc{izacard2022unsuperviseddenseinformationretrieval,
      title={Unsupervised Dense Information Retrieval with Contrastive Learning}, 
      author={Gautier Izacard and Mathilde Caron and Lucas Hosseini and Sebastian Riedel and Piotr Bojanowski and Armand Joulin and Edouard Grave},
      year={2022},
      eprint={2112.09118},
      archivePrefix={arXiv},
      primaryClass={cs.IR},
      url={https://arxiv.org/abs/2112.09118}, 
}

@inproceedings{thorne-etal-2018-FEVER,
    title = "The Fact Extraction and {VER}ification ({FEVER}) Shared Task",
    author = "Thorne, James  and
      Vlachos, Andreas  and
      Cocarascu, Oana  and
      Christodoulopoulos, Christos  and
      Mittal, Arpit",
    editor = "Thorne, James  and
      Vlachos, Andreas  and
      Cocarascu, Oana  and
      Christodoulopoulos, Christos  and
      Mittal, Arpit",
    booktitle = "Proceedings of the First Workshop on Fact Extraction and {VER}ification ({FEVER})",
    month = nov,
    year = "2018",
    address = "Brussels, Belgium",
    publisher = "Association for Computational Linguistics",
    url = "https://aclanthology.org/W18-5501/",
    doi = "10.18653/v1/W18-5501",
    pages = "1--9"
}

@misc{santhanam2022colbertv2effectiveefficientretrieval,
      title={ColBERTv2: Effective and Efficient Retrieval via Lightweight Late Interaction}, 
      author={Keshav Santhanam and Omar Khattab and Jon Saad-Falcon and Christopher Potts and Matei Zaharia},
      year={2022},
      eprint={2112.01488},
      archivePrefix={arXiv},
      primaryClass={cs.IR},
      url={https://arxiv.org/abs/2112.01488}, 
}

@misc{lee2025nvembedimprovedtechniquestraining,
      title={NV-Embed: Improved Techniques for Training LLMs as Generalist Embedding Models}, 
      author={Chankyu Lee and Rajarshi Roy and Mengyao Xu and Jonathan Raiman and Mohammad Shoeybi and Bryan Catanzaro and Wei Ping},
      year={2025},
      eprint={2405.17428},
      archivePrefix={arXiv},
      primaryClass={cs.CL},
      url={https://arxiv.org/abs/2405.17428}, 
}

@misc{muennighoff2025GritLMgenerativerepresentationalinstructiontuning,
      title={Generative Representational Instruction Tuning}, 
      author={Niklas Muennighoff and Hongjin Su and Liang Wang and Nan Yang and Furu Wei and Tao Yu and Amanpreet Singh and Douwe Kiela},
      year={2025},
      eprint={2402.09906},
      archivePrefix={arXiv},
      primaryClass={cs.CL},
      url={https://arxiv.org/abs/2402.09906}, 
}

@misc{wang2024improvingtextembeddingslarge,
      title={Improving Text Embeddings with Large Language Models}, 
      author={Liang Wang and Nan Yang and Xiaolong Huang and Linjun Yang and Rangan Majumder and Furu Wei},
      year={2024},
      eprint={2401.00368},
      archivePrefix={arXiv},
      primaryClass={cs.CL},
      url={https://arxiv.org/abs/2401.00368}, 
}

@misc{trivedi2023ircot,
      title={Interleaving Retrieval with Chain-of-Thought Reasoning for Knowledge-Intensive Multi-Step Questions}, 
      author={Harsh Trivedi and Niranjan Balasubramanian and Tushar Khot and Ashish Sabharwal},
      year={2023},
      eprint={2212.10509},
      archivePrefix={arXiv},
      primaryClass={cs.CL},
      url={https://arxiv.org/abs/2212.10509}, 
}

@misc{asai2023SelfRaglearningretrievegenerate,
      title={Self-RAG: Learning to Retrieve, Generate, and Critique through Self-Reflection}, 
      author={Akari Asai and Zeqiu Wu and Yizhong Wang and Avirup Sil and Hannaneh Hajishirzi},
      year={2023},
      eprint={2310.11511},
      archivePrefix={arXiv},
      primaryClass={cs.CL},
      url={https://arxiv.org/abs/2310.11511}, 
}

@misc{edge2025GraphRAG,
      title={From Local to Global: A Graph RAG Approach to Query-Focused Summarization}, 
      author={Darren Edge and Ha Trinh and Newman Cheng and Joshua Bradley and Alex Chao and Apurva Mody and Steven Truitt and Dasha Metropolitansky and Robert Osazuwa Ness and Jonathan Larson},
      year={2025},
      eprint={2404.16130},
      archivePrefix={arXiv},
      primaryClass={cs.CL},
      url={https://arxiv.org/abs/2404.16130}, 
}

@misc{asl2025fairragfaithfuladaptiveiterative,
      title={FAIR-RAG: Faithful Adaptive Iterative Refinement for Retrieval-Augmented Generation}, 
      author={Mohammad Aghajani Asl and Majid Asgari-Bidhendi and Behrooz Minaei-Bidgoli},
      year={2025},
      eprint={2510.22344},
      archivePrefix={arXiv},
      primaryClass={cs.CL},
      url={https://arxiv.org/abs/2510.22344}, 
}

@misc{nahid2025PRISMagenticretrievalllms,
      title={PRISM: Agentic Retrieval with LLMs for Multi-Hop Question Answering}, 
      author={Md Mahadi Hasan Nahid and Davood Rafiei},
      year={2025},
      eprint={2510.14278},
      archivePrefix={arXiv},
      primaryClass={cs.CL},
      url={https://arxiv.org/abs/2510.14278}, 
}

@article{muennighoff2022mteb,
  author = {Muennighoff, Niklas and Tazi, Nouamane and Magne, Loïc and Reimers, Nils},
  title = {MTEB: Massive Text Embedding Benchmark},
  publisher = {arXiv},
  journal={arXiv preprint arXiv:2210.07316},
  year = {2022},
  url = {https://arxiv.org/abs/2210.07316},
  doi = {10.48550/ARXIV.2210.07316},
}

\clearpage
\appendix
\section{Appendix}
\label{sec:appendix}
\subsection{Implementation Details and Hyperparameters}
\label{app:implementation}
We summarize the core hyperparameters for CatRAG in Table \ref{tab:hyperparameters}. To ensure fair comparison, we maintain the QA prompts established in the HippoRAG 2 benchmark \cite{gutiérrez2025hipporag2}.

\begin{table}[h]
    \centering
    \small
    \caption{\textbf{Hyperparameters for CatRAG.} Note that Synonym Edge weights are dynamic, scaled by their vector similarity, whereas the standard HippoRAG 2 framework uses raw vector similarity.}
    \label{tab:hyperparameters}
    \begin{tabular}{lc}
        \toprule
        \textbf{Parameter} & \textbf{Value} \\
        \midrule
        Synonym Similarity Threshold & $0.8$ \\
        Synonym Edge Weight & $2.0 \times \text{Similarity}$ \\
        PPR Damping Factor ($d$) & $0.5$ \\
        LLM Temperature & $0.0$ \\
        Symbolic Anchor ($\epsilon$) & $0.2$ \\
        Max Seed Nodes for scoring ($N_{seed}$) & $5$ \\
        Max Pruning Edges for scoring ($K_{edge}$) & $15$ \\
        Passage Boost Factor ($\beta$) & $2.5$ \\
        \bottomrule
    \end{tabular}
\end{table}

\noindent \textbf{Dynamic Edge Scoring Schedule.}
To translate the LLM's semantic assessment into topological structure, we employ a tiered projection strategy. We define four distinct semantic tiers—Irrelevant, Weak, High, and Direct—and map the discrete LLM scores $s \in \{0, \dots, 10\}$ to specific weight intervals (Table \ref{tab:mapping_schedule}). This non-linear mapping acts as a high-pass filter, strictly pruning noise (scores $\le 3$) while exponentially amplifying high-confidence evidence paths.

\begin{table}[h]
    \centering
    \small
    \caption{\textbf{LLM Score to Edge Weight Projection.} Discrete relevance scores are mapped to weight intervals. Within the \textit{Weak} and \textit{High} tiers, weights are linearly interpolated.}
    \label{tab:mapping_schedule}
    \begin{tabular}{lccl}
        \toprule
        \textbf{Semantic Tier} & \textbf{LLM Score ($s$)} & \textbf{Output Weight $\phi(s)$} \\
        \midrule
        \textit{Irrelevant} & $0 - 3$ & $0$  \\
        \textit{Weak} & $4 - 6$ & $0.2 - 0.3$ \\
        \textit{High} & $7 - 9$ & $2.0 - 3.0$  \\
        \textit{Direct} & $10$ & $5.0$ \\
        \bottomrule
    \end{tabular}
\end{table}

\clearpage
\section{Prompts}
\label{prompt}
\label{summary_prompt}
\begin{table}[h]
    \centering
    \begin{adjustbox}{max width=\textwidth}
    \begin{tabular}{p{\textwidth}}
    \toprule
        {\bfseries Entity Summarization Prompt (Adaptive Entity Context)} \cr
    \midrule
    \ttfamily
    \small
    --- Task --- \newline
    Generate a concise, entity-focused summary that captures the core identity and key relationships of a given entity based on its associated fact triplets. \newline
    \newline
    --- Instructions --- \newline
    1. **Input Format**: You will receive: \newline
    \hspace*{1em}- A `target\_entity` (the entity being summarized) \newline
    \hspace*{1em}- A `fact\_triplets` list in JSON format containing relationships where this entity appears \newline
    \newline
    2. **Output Requirements**: \newline
    \hspace*{1em}- Focus on the **target entity** as the summary's subject \newline
    \hspace*{1em}- Integrate ALL key relationships from the provided triplets \newline
    \hspace*{1em}- Explain **what the entity is** and **what it connects to** through its relationships \newline
    \hspace*{1em}- Maintain strict coherence and factual accuracy \newline
    \hspace*{1em}- Maximum length: 150 tokens \newline
    \hspace*{1em}- Language: English (preserve proper nouns in original form when needed) \newline
    \newline
    3. **Content Guidelines**: \newline
    \hspace*{1em}- Start with the entity's core identity/type \newline
    \hspace*{1em}- Group related relationships logically (e.g., all professional roles together) \newline
    \hspace*{1em}- Highlight notable connections to other significant entities \newline
    \hspace*{1em}- Avoid listing facts mechanically - synthesize into narrative form \newline
    \newline
    --- Example Structure --- \newline
    [Entity Name] is a [core type/description] known for [key attributes]. It [main relationships/activities] with entities such as [notable connections]... \newline
    \newline
    \textit{[... One in-context learning examples  ...]} \newline
    \newline
    --- Input --- \newline
    Target node: \$\{entity\} \newline
    Fact Triplets: \$\{fact\_triplets\} \\
    \bottomrule
    \end{tabular}
    \end{adjustbox}
    \caption{Prompt for generating entity summaries.}
\end{table}

\clearpage
\begin{table}[h]
    \centering
    \begin{adjustbox}{max width=\textwidth}
    \begin{tabular}{p{1.0\textwidth}}
    \toprule
        \textbf{Knowledge Graph Neighbor Scoring Prompt (Fine-Grained Semantic Probability Alignment)} \\
    \midrule
    \ttfamily
    \small
    You are a knowledge graph reasoning expert. Score neighbor entities (0-10) on their utility for answering a QUERY. \newline
    \newline
    \#\#\# Input Data \newline
    1. A user QUERY. \newline
    2. The CURRENT ENTITY node we are exploring. \newline
    3. A set of RETRIEVED FACTS (trusted evidence). \newline
    4. A list of NEIGHBORS, each with: \newline
    \hspace*{1em}- The specific LINKING TRIPLET(s) connecting the current entity to this neighbor. \newline
    \hspace*{1em}- A short summary of the neighbor information. \newline
    \newline
    \#\#\# Scoring Criteria \newline
    - **10 (Solution):** The neighbor IS the answer or contains it. \newline
    - **7-9 (Bridge):** Critical step in the reasoning chain (e.g., Subject -> Attribute). \newline
    - **4-6 (Weak):** Valid semantic link, but tangential to query intent. \newline
    - **0-3 (Noise):** Irrelevant, generic, or contradicts facts. \newline
    \newline
    \#\#\# Rules \newline
    1. **Trust Facts:** If a neighbor contradicts RETRIEVED FACTS, score 0. \newline
    2. **Output Format:** - `ID (Entity Name): Score` (if Score < 4) \newline
    \hspace*{1em}- `ID (Entity Name): Score | Concise reasoning` (if Score >= 4) \newline
    3. **Constraint:** You must copy the Entity Name exactly as it appears in the input. \newline
    \newline
    \textit{[... Two in-context learning examples ...]} \newline
    \newline
    Output ONE line per neighbor: `ID (Entity Name): Score | (Reasoning if Score >= 4)` \\
    \bottomrule
    \end{tabular}
    \end{adjustbox}
    \caption{The prompt for scoring neighbor nodes.}
\end{table}

\end{document}